\documentclass[journal,twocolumn,11pt]{IEEEtran}
\usepackage{multirow}

\usepackage{cite}
\usepackage{amsmath,amssymb,amsfonts}
\usepackage{algorithmic}
\usepackage{graphicx}
\usepackage{textcomp}
\usepackage{matrixex}
\usepackage{xcolor,soul}
\usepackage{url}
\usepackage{array}
\newtheorem{defn}{Definition}
\def\BibTeX{{\rm B\kern-.05em{\sc i\kern-.025em b}\kern-.08em
    T\kern-.1667em\lower.7ex\hbox{E}\kern-.125emX}}

\begin{document}


\title{On the Gap between Epidemiological Surveillance and Preparedness  }
\author{\IEEEauthorblockN{ 
Svetlana  Yanushkevich, Vlad Shmerko\\
}
\IEEEauthorblockA{
\textit{Biometric Technologies Laboratory, ECE Department}\\ \textit{ University of Calgary, Canada}\\ 
http://www.ucalgary.ca/btlab, syanshk@ucalgary.ca}
}

\markboth{2020}
{...\MakeLowercase{\textit{S. Yanushkevich, V. Shmerko}}:
...}

\maketitle

\thispagestyle{empty}

\begin{abstract}
	Contemporary Epidemiological  Surveillance (ES) relies heavily on data analytics. These analytics are critical input for pandemics preparedness networks; however, this input is not integrated into a form suitable for decision makers or experts in preparedness. A decision  support system (DSS) with Computational Intelligence (CI) tools is required  to bridge the  gap between epidemiological model of evidence and expert group decision. We argue that  such DSS shall be a 		cognitive dynamic system enabling the CI and human expert to work together. The core of such DSS must be based on machine reasoning techniques  such as probabilistic  inference, and shall be capable of estimating risks, reliability and biases in decision making.	
\end{abstract}
			


\maketitle

\section{Introduction}
\label{sec:}


   World Health Organization distinguishes five levels of strategic preparedness for pandemic outbreak \cite{WHO2020}:
    \begin{enumerate}
  \item Community Transmission,
  \item Local Transmission,
  \item Imported cases,
  \item High risk of imported cases,
  \item Preparedness.
\end{enumerate}

 The entity that monitors, reports and develops the policies and strategies to prepare for response  is usually called a Preparedness network. An example of at attempts to create such infrastructure was  a project called the European S-HELP (Securing-Health Emergency Learning Planning, 2014-2017), which, however, was not reported to be materialized into a functioning system.

Epidemiological Surveillance (ES) is  a part of  any local, national or global epidemic preparedness network such as the Global Influenza Surveillance and Response System (GISRS)  \cite{GISRS},  the Communicable Diseases Network Australia (CDNA) and the Commonwealth’s National Notifiable Diseases Surveillance System (NNDSS).
The ES has means to assess a state of epidemic threat in order to introduce suitable countermeasures. In GISRS, for example, the public health decision-making is based on assessing pandemic severity and transmissibility score as guided by Pandemic Severity Assessment Framework \cite{Reed2013}.
Early in a pandemic, when limited data are available, scores are
 rated as low-moderate or moderate-high. As additional data become available,  transmissibility is assigned a value  within a range of 1–5, and severity is varied within the range 1–7. These characterize the potential pandemic impact in relation to previous pandemics or seasonal epidemics. The further assessment of  impact on response and mitigation of an emerging pandemic is not generally supported by any Computational Intelligence (CI) tool.

\begin{figure*}[!ht]
\begin{center}
	\includegraphics[scale=0.7]{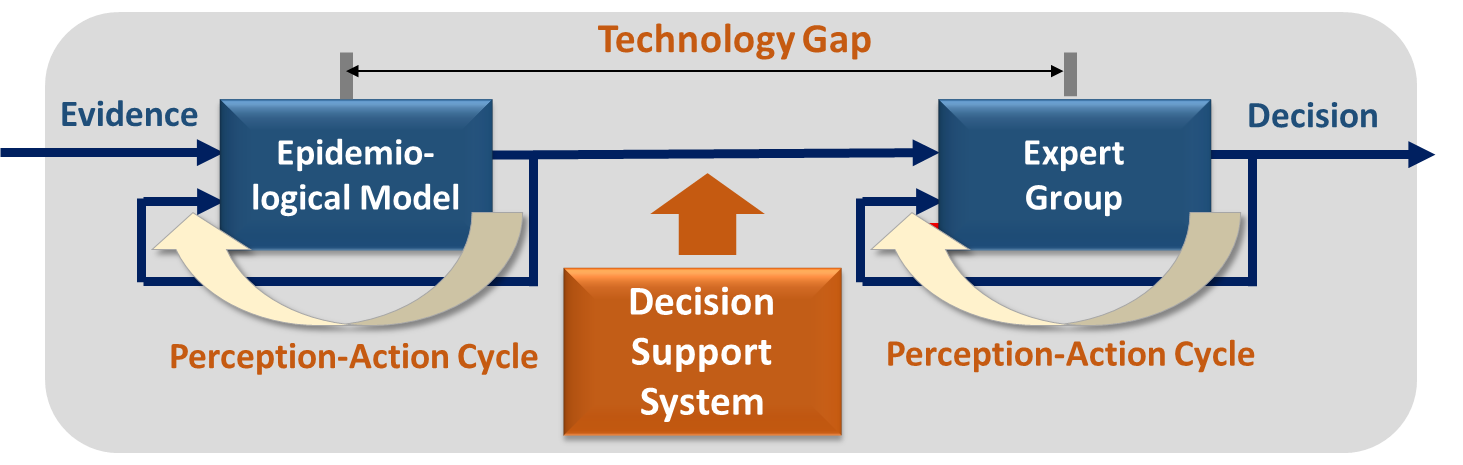}
\end{center}
	\caption{Contemporary ES practice: the result of the epidemic or pandemic modeling, which is based on current evidence and most of the time ignores  multiple variable influencing the predictions, is provided directly to the human users, or a group of experts. This output does not usually include the results of new knowledge inference, or reasoning, to help experts to cope with abundance of data and uncertainty, and project them into the usable measures as risk or bias}
	\label{fig:Gap}
\end{figure*}

In biostatistics, various ES models have been developed and used by statisticians and epidemiologists. However, they generally are  not being translated in the practice of preparedness for pandemic. The ES models can predict the pandemics mortality and morbidity, but do not account for multiple  variables and uncertainties such as reliability of surveyed data or credibility of survey or testing technology. they are not directly presented in the form that enables its use in pandemics response or mitigation.

 It is known that pandemics assessment ``depends on rapid availability of treatment, clinical support, and vaccines''  \cite{Reed2013}, and those variables are not  taken into account in the known statistical ES models. Some  ES models can be ``stratified'' for age, gender of other variables, but does not provide any causal analysis. In other words, no   decision support tool, that would provide meta-analysis and assess risks in preparedness context, is readily available.

COVID-19 outbreak unveiled critical  disadvantages of the existing ES model and  in providing support  for preparedness systems. 
Important lessons learnt from these failures are as follows:
\begin{enumerate}
	\item Outcomes of the ES models need to be further translated to become usable. 
		\item Experts and PN management teams require  computational intelligence (CI) support in decision-making process that is ported into the ES model.
\end{enumerate}
At a system level, these disadvantages manifest themselves
as a technology gap, as illustrated in Fig. \ref{fig:Gap}.

	In terms of a community, city or country preparedness as a whole, much more factors come into consideration in order to implement epidemic preparedness. An Epidemic Preparedness Index (EPI)  developed in \cite{Oppenheim2019} includes 23 indicators grouped into five sub-indexes:
   \begin{enumerate}
  \item Public Health Infrastructure,
  \item Physical Infrastructure,
  \item Institutional Capacity,
  \item Economic Resources,
  \item Public Health Communication. 
\end{enumerate}
This rating  measures the national-level preparedness for major outbreaks of infectious diseases, based on various factors related to the healthcare system.

A DSS to be used in  Epidemic Preparedness system at  community, city or country-level shall include a system of DSSs, each takes  into account the support to the experts that manage various community infrastructures and resources such as the  ones listed in \cite{Oppenheim2019}

\begin{figure*}[!ht]
\begin{center}
\includegraphics[scale=0.7]{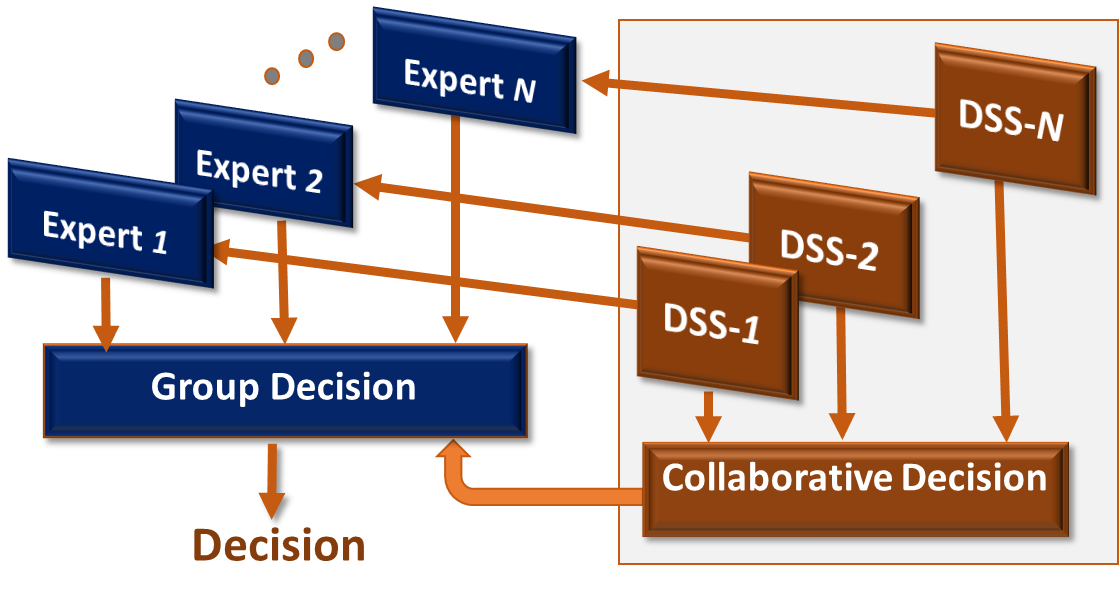}
 \end{center}
 \caption{A framework of the DSS emerging applications: human-machine teaming  using  DSS. Final decision is the consensus of expert's decisions supported by consensus of DSS.}
  \label{fig:Group-DSS}
\end{figure*}

 This approach is based on the concept of group decision making \cite{[Bedford-2013],[Kamisa-2018]} that is illustrated in Fig. \ref{fig:Group-DSS}. Given an evidence  and  $N$ experts, each of them is supported by the DSS; the task to support experts in group decision-making.

 We state that each expert needs a specific-purpose DSS related to their respected field of expertise. 
Given data from the epidemiological model, the output of each DSS is a result of a \emph{dynamic evidential reasoning}. The general principles of building such DSS, as outlined in  this paper, shall be applied to build the aforementioned hierarchy of DSSs for better managing future epidemics and pandemics. 



\section{Contribution}
\label{sec:Contr}

 Uncertainties in developing epidemic/pandemic is equally unavoidable; hence, experts require a CI tool that accounts for uncertainty. The answer lies in usage of probabilistic reasoning, namely,  causal (Bayesian) networks operating using probabilities and enabling knowledge inference based on priors and evidence \cite{[Pearl2019]}. This approach has been  applied to risk assessment in multiple areas of engineering and economics \cite{Fenton-book}, risk profiling \cite{[Yanushkevich-2018]}, identity management \cite{[Yanush-2019-a]}, precision medicine diagnostics \cite{[Vinarti-2019]} and   very recently to  analysis of COVID-19 risks such as fatality and prevalence rates \cite{[Neil2020]}.

 This paper advocates for a concept of a Decision Support System (DSS) for ES, with a CI core based on causal Bayesian networks. We define a DSS as a crucial bridging component to be integrated into the existing  ES systems in order to provide situational awareness and help handle the outbreaks better. 

The DSS concept has been developed for multiple applications, and were once known as ``expert systems''. For example,  paper \cite{[Zachary-1997]}  is still a useful guide for  training experts and DSS design, e.g. automation of reasoning and interpretation strategies is the way to extend experts' abilities to apply their own strategies efficiently.  

 Specific-centered DSS is a  well-identified trend in using the CI in many fields including   the ES. 
Examples of  contemporary DSS are personal health monitoring systems \cite{[Andreu-2016]}, e-coaching for health \cite{[Ochoa-2018]},
 security checkpoints  \cite{[Labati-2016],[Yanush-2019]}, and multi-factor authentication systems \cite{[Roy-2018]}.

Usability of CI tools to support a team of practitioners with or without technical background should be estimated  using various measures  compared to the decision making tools performance.   Risk, reliability, trust and bias  are emerging ``precaution'' measures for the cognitive DSS.  
The performance of the cognitive   DSS is evaluated in various dimensions:
\begin{itemize}
	\item [$-$]  Technical, e.g. prediction  accuracy and throughput  \cite{[Roy-2018]},
	\item [$-$] Social, e.g. trust in CI's reliability \cite{[AI-now-Report-2018],[Danks_2017]},
	\item [$-$] Psychological, e.g., efficiency of human-machine interactions \cite{[Montibeller-2015],{[Hou-2011]}}, and
	\item [$-$] Privacy and security domain, e.g., vulnerability of personal data 	\cite{[Andreou-2017],[Yanush-2019],[Bellovin-2019],[EU-Fundamental-Rights-2019]}. 
\end{itemize}

In our study, we model the DSS as a complex multi-state dynamic system \cite{[Yanush-2019]}. The risk assessment is an essential part of such a system \cite{[Yanush-2019-a]}. The crucial idea of our approach is that risk, reliability, trust and bias should be estimated using reasoning mechanisms.



Some of the risk, reliability and trust projections have been studied in technical systems \cite{[Andreou-2017],[Feng-2014]} and social systems \cite{[Schaupp-2010]}. 

Failure of the existing ES to provide proper COVID-19 outbreak modeling and prediction for preparedness infrastructure revealed significant gaps in  the existing concept, design, deployment, and collaboration of the national and international preparedness networks.  In this study, we \textbf{identify a technological gap} in the ES in both technical and conceptual domains (Fig. \ref{fig:Gap}). Conceptually, the  ES users require a significant cognitive support using   a distributed computational intelligence tools.    
This paper addresses the key research question: \textbf{How to bridge this gap using the DSS concept?}

 To answer this question,  we have chosen the model of a DSS called a \textbf{dynamic cognitive platform}. We follow a well-identified trend in academic discussion on the future generation DSS  \cite{[Lai-2020]}.  Unfortunately, this is an important  but partial solution.

 This paper make further steps and    contributes to practice of  technology gap bridging. The key contribution is twofold:
\begin{enumerate}
\item Development of a   \textbf{reasoning and prediction mechanism, -- the core of a DSS}; for this, a concept of a Bayesian causal network \cite{[Pearl1988]} is used; in particular, recent real-world scenario of COVID-19 was described using a Bayesian network \cite{[Neil2020]}.
\item Development of a \textbf{complete spectrum of the risk and bias measures}, including  ES taxonomy  updating. 
\end{enumerate}

These results are coherent with the solutions to the following related problems:
\begin{enumerate}
\item [$-$] The technology gap ``pillars'' in Fig. \ref{fig:Gap} are Protocol of the ES model and Protocol of DSS. These protocols are different, e.g. spread virus behavior and  conditions of small business operation.   The task is to convert  the ES  protocol  into a DSS specification. Criteria of efficiency of conversion are an acceptance of a given field expert. \textbf{Reasoning mechanism based on causal network intrinsically contains  the protocol conversion}. We demonstrate this phenomenon in our experiments.
\item [$-$] The DSS supports an expert to make decisions under uncertainty in a specific field of an expertise. Specifically, intelligent computations help an expert in better interpretation of uncertainty under chosen precautions. \textbf{The risk and bias  are used in this paper as a  precautions of different kind of uncertainties} related to ES data \cite{[Glossary]},  testing tools, human factors, ES model turning parameters, and artificial intelligent.  
\item [$-$] \textbf{Standardization} at all levels of the ES of preparedness network including the international links is essential in combating epidemic and pandemics. For example, different formats (protocols, standards) are defined as a significant obstacle in Covid-19 pandemic \cite{[Alamo-2020]}.
An example of taxonomy of information uncertainty in military situational awareness is an Admiralty Code used  in NATO standard. We propose to use NATO experience in this area for the pandemics preparedness purposes.
\end{enumerate}  
A DSS  concept suitable for the ES model is proposed in this paper.

\section{Background}
\label{sec:Background}

In the face of future pandemics, challenging problems in preparedness include:

 1) Improvement of the models in order to provide more reliable recommendations that are acceptable for real-world scenarios, and

 2) Mitigation of the “incorrectness” of pandemic models in the decision-making response.

We advocate  for a solution that utilizes the foremost machine intelligence techniques called machine reasoning (knowledge extraction and inference mechanism) that would provide information support to the decision makers.

\subsection{ Risk  taxonomy}

A cognitive DSS is a semi-automated system, which deploys CI to process the data sources and to assess risks or biases; this assessment is submitted to a human operator for the final decision.

The risk, reliability, trust and bias measures are used in ES in simple forms such as `high-risk group', `risk factor', and `systematic difference in the enrollment of participants'  \cite{[Glossary]}. However, experts  expect from the cognitive DSS more detailed assessments of epidemic scenarios because the corpus of  risk terms and definitions is limited. 
For example, syndrome surveillance consists of real-time indicators for disease that allow for early detection. The DSS must support an expert with answers to the following questions:  How risky a given state of ES with respect to collapse of health care resources? Can experts rely on the collected data? What kind of biases can be expected in data collection, algorithmic processing, and the decision making?

\begin{defn}\label{def:Risk}
	\textbf{Risk} is a measure of the extent to which an entity is threatened by a potential circumstance or event, and typically is a function of: (i) the adverse \textbf{impact}, also called \textbf{cost} or magnitude of harm, that would arise if the circumstance or event occurs, and (ii) the \textbf{likelihood} of event occurrence~\cite{[NIST-2017]}. $\blacksquare$
\end{defn}
For example, in automated decision making, and  in our study, the \texttt{Risk} is defined as a function   $F$ of cost (or consequences) of a circumstance or event and its occurrence probability: 

\begin{footnotesize}
\begin{center}
	\definecolor{light}{gray}{.9}
	\colorbox{light}{
		\begin{parbox}[h]{0.95\linewidth} {
			\vspace{-2mm}
			\begin{center}
				\[\texttt{Risk} = F(\texttt{Cost},\ \texttt{Probability})\]
	\end{center}} \end{parbox}}
\end{center}
\end{footnotesize}

\begin{defn}\label{def:Bias}
\textbf{Bias} in the cognitive DSS refers to the tendency of an assessment process to systematically over- or under-estimate the value of a population parameter.~$\blacksquare$
\end{defn}
For example, in the context of detecting or testing for  an infectious disease, the bias is related to the sampling approaches (e.g. tests are performed on a proportion of cases only),  sampling methodology (systemic or random), and chosen testing procedures or devices \cite{WHO}.

While all these biases are different, they are probabilistic in nature because the evidence and information gathered to make a decision is always incomplete, often inconclusive, frequently ambiguous, commonly dissonant, conflicting, and has various degrees of believability. 	Identifying and mitigating bias is essential for assessing decision risks and AI biases  \cite{[AI-now-Report-2018],[Gates-2018],[Lai-2020]}.


%
%


 For the cognitive DSS to operate effectively on the human's behalf, the system may need to use confidential or sensitive information of the users such as  personal contact information \cite{[Clavell-2017]}. The users and the operators must be confident that the cognitive DSS will do what they ask, and only what they ask. Human acceptance of the cognitive DSS technology is determined by the combination of the bias factors  and risk factors  \cite{[Anand-2013],[Feng-2014]}. The contributing factors include belief, confidence, experience, certainty, reliability, availability, competence, credibility, completeness, and cooperation \cite{[Zhang-2014],[Cho-2015]}. 
 In our approach, the causal inference platform calculates various uncertainty measures  \cite{[Yanushkevich-2018]} in risk and bias assessment scenarios.


\subsection{Risk and Decision Reliability interpretation}\label{sec:Admiralty-Code}

Risk in decision making process manifests itself  in various ways, such as the reliability of information  sources (e.g. infection survey data) and the credibility of the information (e.g. testing procedure accuracy):

\begin{footnotesize}
\begin{center}
	\definecolor{light}{gray}{.9}
	\colorbox{light}{
		\begin{parbox}[h]{0.95\linewidth} {
			\vspace{-2mm}
			\begin{center}
				$\underbrace{\texttt{Data~Reliability}}_{Quality}~{\Leftrightarrow}
				\left\{\hspace{-0.2cm}
				\begin{array}{c}
				\texttt{Risk} \\
				\texttt{Trust} \\
				\texttt{Bias} \\
				\end{array}\hspace{-0.2cm}
				\right\}{\Leftrightarrow}~
				\underbrace{\texttt{Test~Credibility}}_{Reputation}$\\
	\end{center}} \end{parbox}}
\end{center}
\end{footnotesize}

This relationship can be represented as follows:
\begin{itemize}
	\item [$(a)$]\emph{Source reliability} as the quality of being reliable, or trustworthy, is related to 1) risk as a function of potential adverse impact and the likelihood of occurrence, 2) the confidence in quality,  and 3) bias as systematic over- or under-assessment of the parameter of interest.
	\item [$(b)$] \emph{Information credibility} as the reputation impacting one's ability to be believed. In ES, in particular, it can be associated with the credibility of infection testing technology.
\end{itemize}

\begin{figure*}[!ht]
\begin{center}
	\includegraphics[scale=0.8]{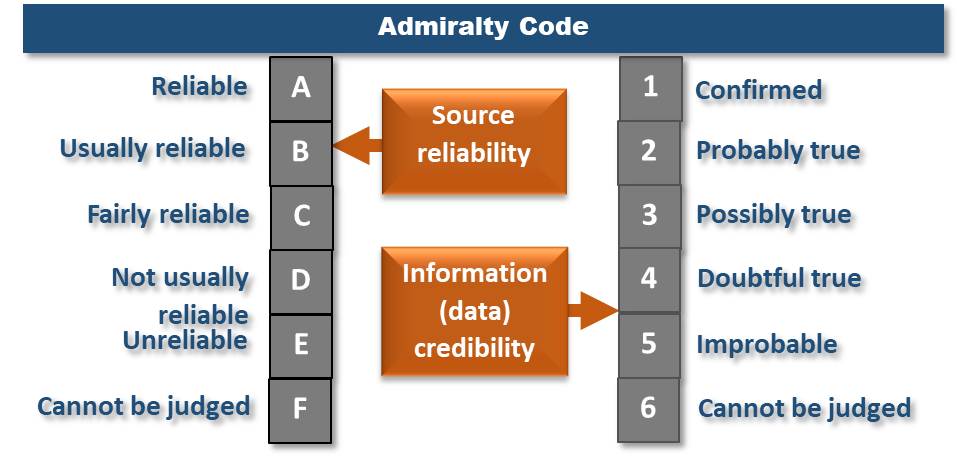}
\end{center}
	\caption{Manifestation of the R-T-B via assessments of reliability of source and  credibility of information using the Admiralty Code.}
	\label{fig:AdmiraltyCode}
\end{figure*}

These metrics of uncertainty closely resemble the ones known as the Admiralty Code defined in the NATO Standardization Agreements such as STANAG 2022 and STANAG 2511 \cite{[Blasch-2013],[Admiralty_Code_2012],[U.S._Army_Field_Manual]} (Fig. \ref{fig:AdmiraltyCode}).  
NATO uses the Admiralty Code to resolve conflicting scenarios in human-human, human-machine, and machine-machine interactions. The reliability of the DSS can be increased by using more reliable sources and creditable information, or it can be diminished due to lowered reliability of the source or/and credibility of the information. For example, scenario $C4$ in Figure \ref{fig:AdmiraltyCode} is composed as the source reliability \texttt{C$\equiv$\ <Fairly Reliable>} and information credibility \texttt{4$\equiv$\ <Doubtful true>}.   In this context, risk can be expressed in terms of the fair reliability of data sources while doubtful credibility of information.

 In this study, we argue that  similar standards shall be developed in the ES practice. Definition of measure of uncertainty in the ES will allow to formalize the reasoning upon uncertainty, and define the variables  to be included in causal networks, and the associated probabilities. In addition, various scenarios developing in epidemic situations, can be characterized by  various levels of uncertainty.

\begin{figure*}[!ht]
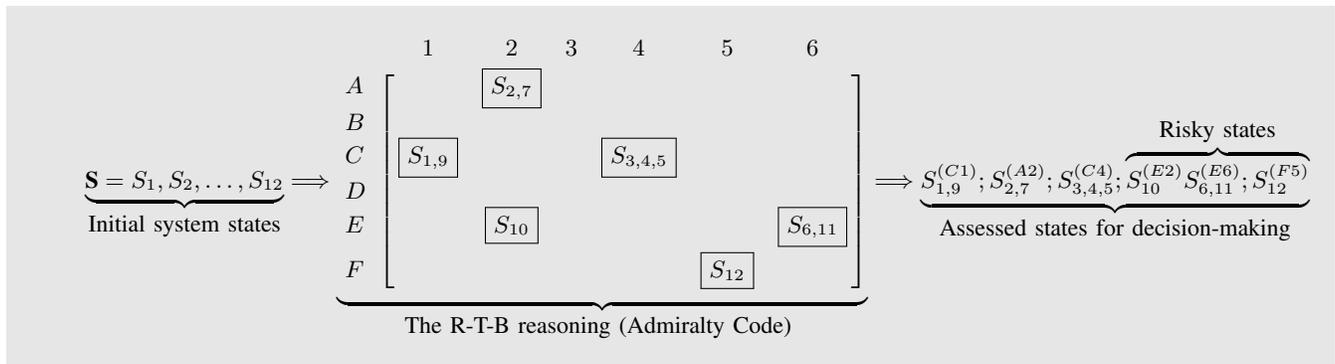


\begin{footnotesize}
\begin{center}
	\definecolor{light}{gray}{.9}
	\colorbox{light}{
		\begin{parbox}[h]{0.95\linewidth} {
			\vspace{-2mm}
			\begin{center}
				\begin{footnotesize}
\begin{eqnarray*}
&&\underbrace{\textbf{S}=S_{1},S_{2},\ldots ,S_{12}}_{\text{\footnotesize Initial system states}}
\Longrightarrow
\underbrace{
\bordermatrixex{
           & \mbox{$1$} & \mbox{$2$} & \mbox{$3$} & \mbox{$4$}& \mbox{$5$}& \mbox{$6$}\cr \noalign{
\vskip 2pt}
\mbox{~$A$}&       &\fbox{$S_{2,7}$}   &&   &   &\cr \noalign{
\vskip 2pt}
\mbox{~$B$}&       &      &   &   &&\cr \noalign{
\vskip 2pt}
\mbox{~$C$}&\fbox{$S_{1,9}$}&      &   &\fbox{$S_{3,4,5}$}&   &\cr \noalign{
\vskip 2pt}
\mbox{~$D$}&    &      &   &   &   &\cr \noalign{
\vskip 2pt}
\mbox{~$E$}&       &\fbox{$S_{10}$}&   &   &   &\fbox{$S_{6,11}$}\cr \noalign{\vskip 2pt}
\mbox{~$F$}&       &      &  &   &\fbox{$S_{12}$}   &\cr \noalign{
\vskip 2pt}
}}_{\text{\footnotesize The R-T-B reasoning (Admiralty Code) }}
\Longrightarrow 
\underbrace{
S_{1,9}^{(C1)};
S_{2,7}^{(A2)};
S_{3,4,5}^{(C4)};
\overbrace{S_{10}^{(E2)}
S_{6,11}^{(E6)};
S_{12}^{(F5)}}^{\text{\footnotesize Risky states}}
}_{\text{\footnotesize Assessed states for decision-making}}
\end{eqnarray*}
\end{footnotesize}
	\end{center}} \end{parbox}}
\end{center}
\end{footnotesize}
\caption{Example of the risk assessment of the epidemic situation. Let the ES be represented by a set of a system states $\textbf{S}=S_i,~i=1,2, \ldots , 12.$ The DSS primary task is the risk reasoning about these states using available resources defined using the Admiralty code (Table \ref{tab:Admiralty-Code}). The result is a set of system states provided to the experts to support their decision making.  }
\label{fig:Admiralty-Code}
\end{figure*}

Fig. \ref{fig:Admiralty-Code} explains the DSS mechanism for assessing different scenarios that are called the \emph{system states}. For example, consider the states $\{S_{1}, S_{2},\ldots , S_{12}\}$ of the ES. The DSS analysis accordingly to the Admiralty Code (Figure \ref{fig:AdmiraltyCode}) results in the following decision-making landscape:

\begin{footnotesize}
\begin{center}
	\definecolor{light}{gray}{.9}
	\colorbox{light}{
		\begin{parbox}[h]{0.95\linewidth} {
			\vspace{-2mm}
			\begin{center}
		\begin{itemize}
	\item [$-$] States $S_{1}$ and $S_{9}$; $S_{2}$ and $S_{7}$; $S_{3},S_{4}$, and $S_{5}$  can be used for decision-making; 
		\item [$-$] States $S_{10}$; $S_{6}$ and $S_{11}$; and $S_{12}$ results in high-risk decisions.
			\end{itemize}
				\end{center}} \end{parbox}}
\end{center}
\end{footnotesize}



\section{Cognitive DSS for ES}
\label{sec:Cogn}

Decision making by human experts vary in knowledge structure, self-interest and growth background, thus the emergence of preference conflict is inevitable. In order to ensure the effectiveness of emergency decision making support, it is imperative to construct a consensus process to reduce and remove preference conflict prior to decision making. This process aligns well with the concept of cognitive DSS.

\subsection{Elements of a cognitive system}

A cognitive DSS  for ES  support is  a complex dynamic system  with the following elements  of a {cognitive} system \cite{[Haykin-2012]}:
\begin{enumerate}
	\item []\hspace{-9mm} \emph{Perception-action cycle} that enables information gain \cite{hou2014intelligent};
	\item []\hspace{-9mm} \emph{Memory} distributed across the entire system (personal data are collected in the physical and virtual world); 
	\item []\hspace{-9mm} \emph{Attention} is driven by memory to prioritize the allocation of available resources; and 
	\item []\hspace{-9mm} \emph{Intelligence} is driven by perception, memories, and attention; its function is to enable the control and decision-making mechanism to help identify intelligent choices. These cognitive elements are distributed throughout the system in a multi-state perception-action cycle~\cite{[Yanush-2019],[Yanush-2019-a]}. 
\end{enumerate}


In most generic sense, the perception-action cycle of a cognitive dynamic system consists of an actuator, an  environment, and a perceptor that are embodied in a feedback loop \cite{[Haykin-2012]}: 
\begin{itemize}
	\item [$-$] The actuator is represented by expert team  and their DSS; they make decisions and actions using available perceived information. 
	\item [$-$] The environment is represented by an epidemiological evidence (real or modeled).
\end{itemize}


\subsection{DSS Design Flow}

The DSS  design process including performance evaluation is shown in 
\begin{figure*}[!ht]
 \begin{center}
\includegraphics[scale=0.7]{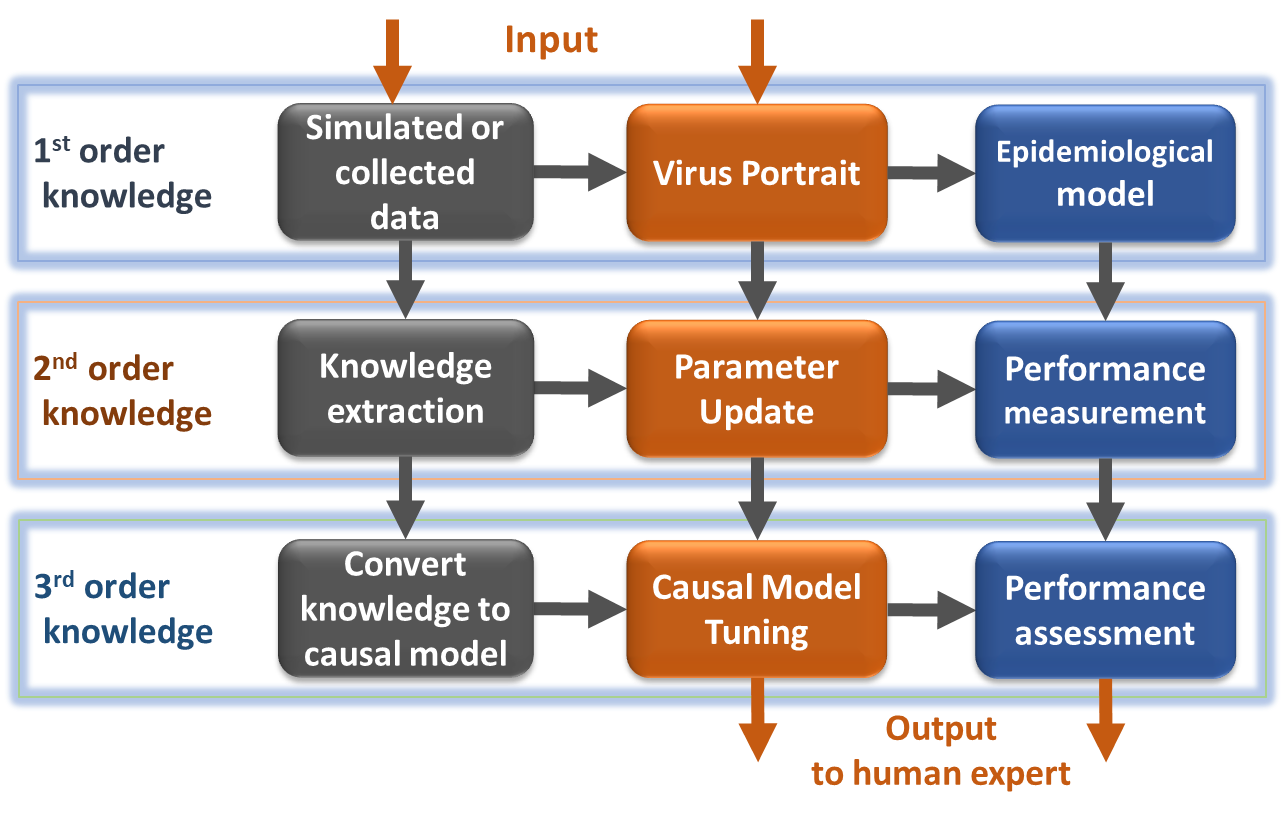}
 \end{center}
\caption{DSS design flow: the data is collected in order to extract knowledge about causality of variables, and convert the uncertainty of data into the Conditional probabilities assigned to the nodes of the casual model; the probabilities are assigned using the procedure of interpreting the discrete values (such as symptomatic or asymptomatic), and discretization of continuous values such as distribution of the numbers of infected based on the  modified SIR models. }  
  \label{fig:Flow}
\end{figure*}

Figure \ref{fig:Flow}. The important parts of this process include system ``precaution'' measurement of the models used for measurement of uncertainty, as well as the DSS precaution measurement in terms of risks and biases with respect to the recommended decisions.

%
%


\subsection{Fundamental DSS operations}

In our work, given a certain causal network platform, the reasoning operations are defined as follows:


\emph{1) State assessment,} such as the risk of the data source being unreliable. In modeling, risk is represented by a corresponding probability distribution function. 

\emph{2) Causal analysis} is based on the ``cause-effect'' paradigm. In particular, Granger causality analysis is an advanced tool for this purpose \cite{[Spirtes-2016]}.

\emph{3) Risk reliability, trust and bias learning.} In \cite{[Rettinger-2008]}, an approach to learning the trust model has been proposed. 

\emph{4) Risk, trust and bias propagation.} The risk propagation problem was studied, in particular, as a multi-echelon supply chain problem in \cite{[Ojha-2018]}. In \cite{[Vydiswaran-2011]}, the trustworthiness is propagated through a three-layer model consisting of a source layer, evidence layer, and claim layer. The quality of evidence was used to compute the trustworthiness of sources. Trust can be propagated through a subjective network \cite{[Ivanovska-2015]}.


\emph{5) Reasoning} is the ability to form an intelligent conclusion or judgment using the evidence. Causal reasoning is a judgment under uncertainty based on a causal network \cite{[Pearl2019]}. 

\emph{6) Risk, reliability, trust  and bias adjustment (or calibration)} aims to improve the confidence of the risk assessment. For example, negative testing in a symptomatic patient may result in a high risk indicator, but can be later adjusted using additional testing.

\emph{7) Risk, reliability, trust and bias mitigation.} In most scenarios, the result of intelligent risk processing is a reduction in risk.  Risk is lowered via mitigation measures and periodical re-assessment in ongoing screening processes.

\emph{8) Risk  prediction.} In complex systems, meta-recognition, meta-learning, and meta-analysis can be used to predict the overall success (correct assessment of the risk) or failure (incorrect one). The most valuable information for such risk assessment is in the ``tails'' of the probabilistic distributions \cite{[Davison-Huser-2015],[Stehlik-2010]}. 

\section{Reasoning mechanism}\label{sec:Reas}

\subsection{Causal network}

A causal network is a directed acyclic graph where each node denotes a unique random variable. A directed edge from node \(n_1\) to node \(n_2\) indicates that the value that is attained by \(n_1\) has a direct causal influence on the value that is attained by \(n_2\). Uncertainty inference requires data structures that will be referred to as \emph{Conditional Uncertainty Tables} (CUTs). A CUT is assigned to each node in the causal network. Given a node \(n\), the CUT assigned to \(n\) is a table that is indexed by all possible value assignments to the parents of \(n\). Each entry of the table is a conditional ``uncertainty model'' that varies according to the choice of the uncertainty metric.
	
Analysis of a causal network is out of the scope of this paper. However, we introduce in this paper the systematic criteria for choosing the appropriate computational tools. In addition, some details are clarified in our experimental study. 

\subsection{Types of causal networks}

Thee are multiple approaches to perform case-and-effect analysis under uncertainty using the causal network structure. Depending on the type of uncertainty measure, they causal networks can be divided to  \cite{[Rohmer-2020]}:

 Bayesian  \cite{[Pearl2019]},
 Interval \cite{[DeCampos1994]};
 Imprecise   \cite{[Coolen-2011]}; 
 Credal   \cite{[Cozman-2000]}; 
 Dempster-Shafer  \cite{[Simona-2008]};
 Fuzzy  \cite{[Baldwin-2003]}; and 
 Subjective    \cite{[Ivanovska-2015]}.

The type of a causal network shall be chosen given the DSS model and a specific scenario. The choice depends on the CUT as a carrier of \emph{primary} knowledge and as appropriate to the scenario.
There were several attempts to provide researchers with the \emph{``Guidelines''} for choosing the best causal network platform based on the CUT. 
	Comparison of causal computational platforms for modeling various systems is a useful strategy, such as Dempster-Shafer vs. credal networks \cite{[Misuri-2018]}, and Bayesian vs. interval vs. Dempster-Shafer vs. fuzzy networks \cite{[Yanushkevich-2018],[Yanush-2019-a]}. 

\subsection{Bayesian causal network}

Motivation  of choosing
Bayesian causal networks is driven by the following:

\begin{itemize}
	\item [$-$] a DSS 
 can be described in  \emph{causal} form using  \emph{cause-and-effect analysis}.
		\item [$-$] Bayesian (probabilistic) interpretation of
        uncertainty provides an acceptable reliability for
        decision-making. 
			\end{itemize}

 The Bayesian decision-making
is based on evaluation of a
  \emph{prior} probability given a \emph{posterior} probability and
 \emph{likelihood} (event happening given some history of previous events)

\begin{footnotesize}
\begin{center}
	\definecolor{light}{gray}{.9}
	\colorbox{light}{
		\begin{parbox}[h]{0.95\linewidth} {
			\vspace{-2mm}
				\begin{center}
\[\overbrace{P(\texttt{Hypothesis|Data})}^{Prior} = \]
\[\overbrace{P(\texttt{Data|Hypothesis})}^{Likelihood}\times \overbrace{P(\texttt{Hypothesis})}^{Posterior}\]
				\end{center}} \end{parbox}}
\end{center}
\end{footnotesize}

Let  the nodes of a graph represent  random variables $X=\{x_1,
\ldots , x_m\}$ and links between the nodes represent direct
causal dependencies.  A Bayesian causal network is
based on a  \emph{factored} representation of joint probability
distributions in the form

\begin{footnotesize}
\begin{center}
	\definecolor{light}{gray}{.9}
	\colorbox{light}{
		\begin{parbox}[h]{0.95\linewidth} {
			\vspace{-2mm}
					\begin{eqnarray*}
P(X)=\overbrace{\prod_{i=1}^{m}P(x_i|\overbrace{\texttt{Par}(x_i)}^{Nodes})}^{Factorization}
\end{eqnarray*}
			}	 \end{parbox}}
\end{center}
\end{footnotesize}

where $\texttt{Par}(x_i)$ denotes a set of parent nodes of the
random variable $x_i$. The nodes outside
$\texttt{Par}(x_i)$ are conditionally independent of $x_i$. Hence,
the Bayesian network has a structural part reflecting causal
relationships, and a probability part reflecting the strengths of
the relationships.
  Factoring techniques have been applied to the construction of
  the Bayesian network.

 The posterior
probability of  $A$ is called the \emph{belief} for $A$,
$\texttt{Bel}(A)$. The probability $P(a|b)$ is called the
\emph{likelihood} of $b$ given $a$ and  is denoted $L(b|a)$.

\subsection{Reasoning for infection outbreak and impact prediction}

The presence of epidemic/pandemic  uncertainties is equally unavoidable; hence, experts require a CI approach that accounts for uncertainty. Probabilistic reasoning on  causal (Bayesian) networks enables knowledge inference based on priors and evidence  has been  applied to  diagnostics for precision medicine \cite{[Vinarti-2019]}. 

Most recently, COVID-19 test-specific risk analysis was performed in \cite{[Neil2020]}: the Bayesian inference was applied to learn the proportion of population with or without symptoms from observations of those tested along with observations about testing accuracy.

 \begin{figure*}[!ht]
\begin{center}
\includegraphics[scale=0.6]{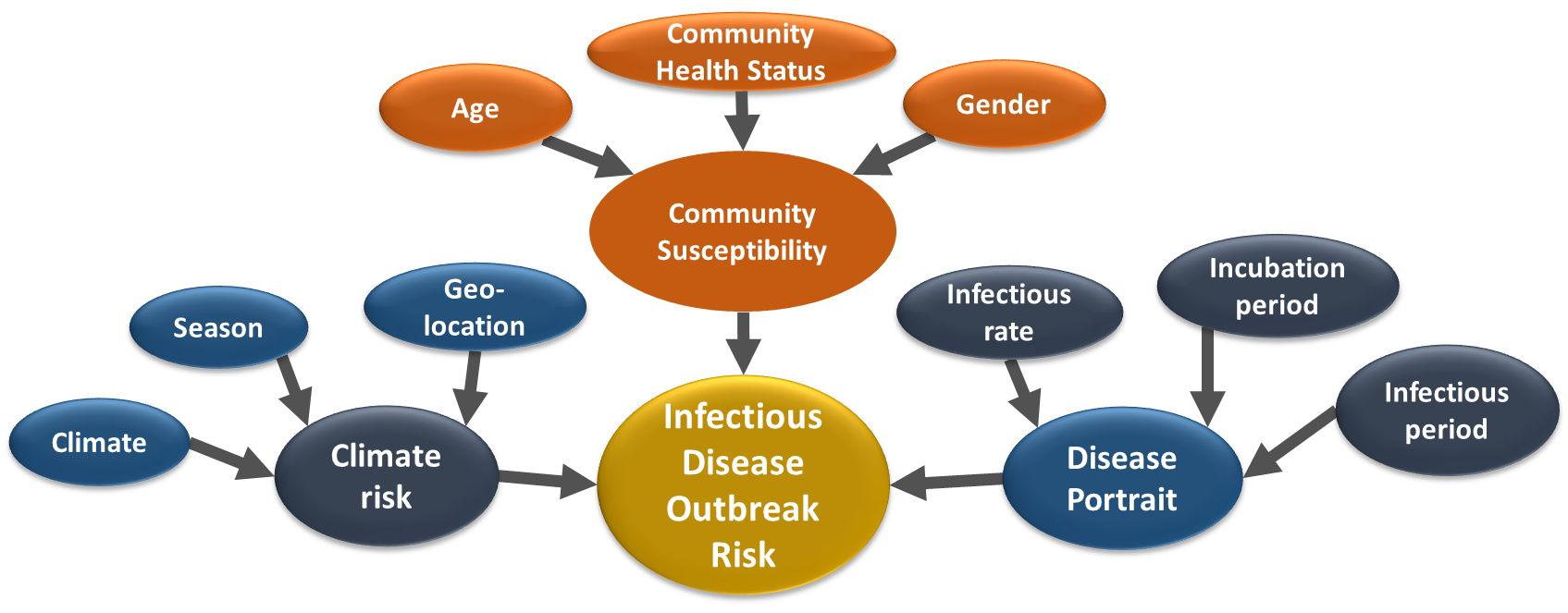}
 \end{center}
 \caption{A simplified causal network used in the core of the  DSS for epidemic preparedness. The nodes represent the variables influencing the decision about the risk of outbreak.}
  \label{fig:BN-prepar}
\end{figure*}

An example of a simplified causal network that can serve as a framework for  risk inference in the context of preparedness is shown in Figure \ref{fig:BN-prepar}. The factors affecting the risk can be evaluated and inferred for various considered scenario. The derivatives such as Preparedness Index can be derived in a similar fashion.

Note that this is only a fragment of a causal network that may be used in the DSS to assess risks related to outbreak itself. 
A DSS to be used in  Epidemic Preparedness system  shall include a system of DSSs. Each expert needs a specific-purpose DSS related to their respected field of expertise. 
Given data from the epidemiological model, the output of each DSS is a result of a \emph{dynamic evidential reasoning}. The general principles of building such DSS, as outlined in  this paper, shall be applied to build the aforementioned hierarchy of DSSs for better managing future epidemics and pandemics.

\section{Conclusions and Future Work}\label{sec:Concl}

 As depicted in Figure \ref{fig:Gap}, there is a technology  gap between the imperfect ES model and human expert's limitations to handle uncertainty  provided by the model while striving to make reliable decisions. Current efforts  regarding  epidemiological models   focus on the following:
 \begin{itemize}
	\item [$-$] Increase \textbf{information gain} about the state of epidemiological threat using perception-action mechanisms.
	\item [$-$] Improve an \textbf{approximation} of joint probabilistic distributions of epidemiological factors; causal networks suggest such possibilities including proactive possibilities.
	\item [$-$]  Create a \textbf{library} of predictive behavior of  virus based on genome studies and deep learning methods \cite{GISAID}. 
	\end{itemize}


 We  propose a general DSS model as a cognitive dynamical system with embedded reasoning mechanism using causal Bayesian network. Additional benefit of this approach is that there is no need in special tools for converting outcome of the model into recommended decisions. 

An \textbf{open applied problem} addresses a ``rational''  partitioning of the model outcome into an ensemble of causal networks. This is because in  a preparedness network, a group of experts from different fields aim to make a decision and come to a certain consensus. Each expert needs a decision support in the respective area such as transportation, hospital readiness, health care, educational institutions, police, counter cyber attack, counter bioterrorism,  etc.  The DSS and human expert's decisions are in causal relations,  they are correlated, and conflicting. This is the field of a group decision making  \cite{[Bedford-2013],[Neville-2015],[Kamisa-2018]}.


\vspace{10mm}




\end{document}